\documentclass{article}

\usepackage{graphicx,caption,subcaption}
\usepackage{xcolor,times,authblk,booktabs,multirow}
\usepackage{array,color,bm,ifthen,tabu,colortbl,dblfloatfix}
\usepackage[sort]{natbib}
\usepackage{algorithm,algorithmic,amsmath,amssymb}
\usepackage{hyperref}

\usepackage[accepted]{icml2017}

\icmltitlerunning{Curiosity-driven Exploration by Self-supervised Prediction}

\begin{document}

\twocolumn[
\icmltitle{Curiosity-driven Exploration by Self-supervised Prediction}

\begin{icmlauthorlist}
\icmlauthor{Deepak Pathak}{ucb}
\icmlauthor{Pulkit Agrawal}{ucb}
\icmlauthor{Alexei A. Efros}{ucb}
\icmlauthor{Trevor Darrell}{ucb}
\end{icmlauthorlist}
\icmlaffiliation{ucb}{University of California, Berkeley}
\icmlcorrespondingauthor{Deepak Pathak}{pathak@berkeley.edu}
\icmlkeywords{curiosity, exploration, reinforcement learning, no reward, self-supervision}

\vskip 0.3in
]
\printAffiliationsAndNotice{}

\begin{abstract}
In many real-world scenarios, rewards extrinsic to the agent are extremely sparse, or absent altogether. In such cases, curiosity can serve as an intrinsic reward signal to enable the agent to explore its environment and learn skills that might be useful later in its life. We formulate curiosity as the error in an agent's ability to predict the consequence of its own actions in a visual feature space learned by a self-supervised inverse dynamics model. Our formulation scales to high-dimensional continuous state spaces like images, bypasses the difficulties of directly predicting pixels, and, critically, ignores the aspects of the environment that cannot affect the agent. The proposed approach is evaluated in two environments: {\em VizDoom} and {\em Super Mario Bros}. Three broad settings are investigated: 1) sparse extrinsic reward, where curiosity allows for far fewer interactions with the environment to reach the goal; 2) exploration with no extrinsic reward, where curiosity pushes the agent to explore more efficiently; and 3) generalization to unseen scenarios (e.g. new levels of the same game) where the knowledge gained from earlier experience helps the agent explore new places much faster than starting from scratch.
\end{abstract}

\section{Introduction}
\label{sec:introduction}
Reinforcement learning algorithms aim at learning policies for achieving target tasks by maximizing rewards provided by the environment. In some scenarios, these rewards are supplied to the agent continuously, e.g. the running score in an Atari game~\cite{mnih2015human}, or the distance between a robot arm and an object in a reaching task~\cite{lillicrap2015continuous}. However, in many real-world scenarios, rewards extrinsic to the agent are extremely sparse or missing altogether, and it is not possible to construct a shaped reward function. This is a problem as the agent receives reinforcement for updating its policy only if it succeeds in reaching a pre-specified goal state. Hoping to stumble into a goal state by chance (i.e. random exploration) is likely to be futile for all but the simplest of environments.

\begin{figure}
    \centering
    \begin{subfigure}[b]{0.48\linewidth}
        \includegraphics[width=\linewidth]{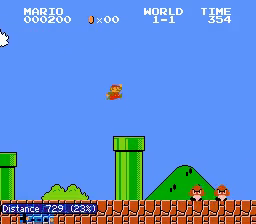}
        \caption{learn to explore in Level-1}
    \end{subfigure}
    ~
    \begin{subfigure}[b]{0.48\linewidth}
        \includegraphics[width=\linewidth]{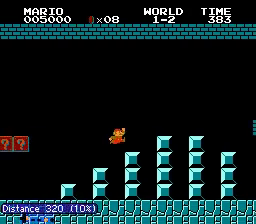}
        \caption{explore faster in Level-2}
    \end{subfigure}
    \vspace{-.12in}
    \caption{Discovering how to play {\em Super Mario Bros} {\bf without rewards}. (a) Using only curiosity-driven exploration, the agent makes significant progress in Level-1. (b) The gained knowledge helps the agent explore subsequent levels much faster than when starting from scratch. Watch the video at \url{http://pathak22.github.io/noreward-rl/}}
    \vspace{-.15in}
    \label{fig:mario_snaps}
\end{figure}

As human agents, we are accustomed to operating with rewards that are so sparse that we only experience them once or twice in a lifetime, if at all. To a three-year-old enjoying a sunny Sunday afternoon on a playground, most trappings of modern life -- college, good job, a house, a family -- are so far into the future, they provide no useful reinforcement signal. Yet, the three-year-old has no trouble entertaining herself in that playground using what psychologists call intrinsic motivation~\cite{Ryan2000} or curiosity~\cite{CuriosityMotivation}.
Motivation/curiosity have been used to explain the need to explore the environment and discover novel states. The French word {\em fl\^aneur} perfectly captures the notion of a
curiosity-driven observer,
the ``deliberately aimless pedestrian, unencumbered by any obligation or sense of urgency'' (Cornelia Otis Skinner). More generally, curiosity is a way of learning new skills which might come handy for pursuing rewards in the future.

Similarly, in reinforcement learning, intrinsic motivation/rewards become critical whenever extrinsic rewards are sparse. Most formulations of intrinsic reward can be grouped into two broad classes: 1) encourage the agent to explore ``novel'' states~\cite{bellemare2016unifying, poupart2006analytic, lopes2012exploration} or, 2) encourage the agent to perform actions that reduce the error/uncertainty in the agent's ability to predict the consequence of its own actions (i.e. its knowledge about the environment)~\cite{schmidhuber1991possibility, schmidhuber2010formal, singh2005intrinsically, mohamed2015variational, stadie2015incentivizing, houthooft2016vime}.

Measuring ``novelty'' requires a statistical model of the distribution of the environmental states, whereas measuring prediction error/uncertainty requires building a model of environmental dynamics that predicts the next state ($s_{t+1}$) given the current state ($s_{t}$) and the action ($a_{t}$) executed at time $t$. Both these models are hard to build in high-dimensional continuous state spaces such as images. An additional challenge lies in dealing with the stochasticity of the agent-environment system, both due to the noise in the agent's actuation, which causes its end-effectors to move in a stochastic manner, and, more fundamentally, due to the inherent stochasticity in the environment. To give the example from \cite{schmidhuber2010formal}, if the agent receiving images as state inputs is observing a television screen displaying white noise, every state will be novel and it would be impossible to predict the value of any pixel in the future. Other examples of such stochasticity include appearance changes due to shadows from other moving entities, presence of distractor objects, or other agents in the environment whose motion is not only hard to predict but is also irrelevant to the agent's goals. Somewhat different, but related, is the challenge of generalization across physically (and perhaps also visually) distinct but functionally similar parts of an environment, which is crucial for large-scale problems.
One proposed solution to all these problems is to only reward the agent when it encounters states that are hard to predict but are ``learnable''~\cite{schmidhuber1991possibility}. However, estimating learnability is a non-trivial problem~\cite{lopes2012exploration}.

This work belongs to the broad category of methods that generate an intrinsic reward signal based on how hard it is for the agent to predict the consequences of its own actions, {\em i.e.} predict the next state given the current state and the executed action. However, we manage to escape most pitfalls of previous prediction approaches with the following key insight: we only predict those changes in the environment that could possibly be due to the actions of our agent or affect the agent, and ignore the rest. That is, instead of making predictions in the raw sensory space (e.g. pixels), we transform the sensory input into a feature space where only the information relevant to the action performed by the agent is represented. We learn this feature space using self-supervision -- training a neural network on a proxy inverse dynamics task of predicting the agent's action given its current and next states. Since the neural network is only required to predict the action, it has no incentive to represent within its feature embedding space the factors of variation in the environment that do not affect the agent itself. We then use this feature space to train a forward dynamics model that predicts the feature representation of the next state, given the feature representation of the current state and the action. We provide the prediction error of the forward dynamics model to the agent as an intrinsic reward to encourage its curiosity.

The role of curiosity has been widely studied in the context of solving tasks with sparse rewards. In our opinion, curiosity has two other fundamental uses.
Curiosity helps an agent explore its environment in the quest for new knowledge (a desirable characteristic of exploratory behavior is that it should improve as the agent gains more knowledge). Further, curiosity is a mechanism for an agent to learn skills that might be helpful in future scenarios. In this paper, we evaluate the effectiveness of our curiosity formulation in all three of these roles.

We first compare the performance of an A3C agent~\cite{mnih2016asynchronous} with and without the curiosity signal on 3-D navigation tasks with sparse extrinsic reward in the {\em VizDoom} environment. We show that a curiosity-driven intrinsic reward is crucial in accomplishing these tasks (see Section~\ref{sec:sparse}). Next, we show that even in the absence of any extrinsic rewards, a curious agent learns good exploration policies. For instance, an agent trained only with curiosity as its reward is able to cross a significant portion of Level-1 in {\em Super Mario Bros}. Similarly in {\em VizDoom}, the agent learns to walk intelligently along the corridors instead of bumping into walls or getting stuck in corners (see Section~\ref{sec:noreward}). A question that naturally follows is whether the learned exploratory behavior is specific to the physical space that the agent trained itself on, or if it enables the agent to perform better in unseen scenarios too? We show that the exploration policy learned in the first level of {\em Mario} helps the agent explore subsequent levels faster (shown in Figure~\ref{fig:mario_snaps}), while the intelligent walking behavior learned by the curious {\em VizDoom} agent transfers to a completely new map with new textures (see Section ~\ref{sec:generalization}). These results suggest that the proposed method enables an agent to learn generalizable skills even in the absence of an explicit goal.

\begin{figure*}[t]
\centering
\includegraphics[width=0.9\linewidth]{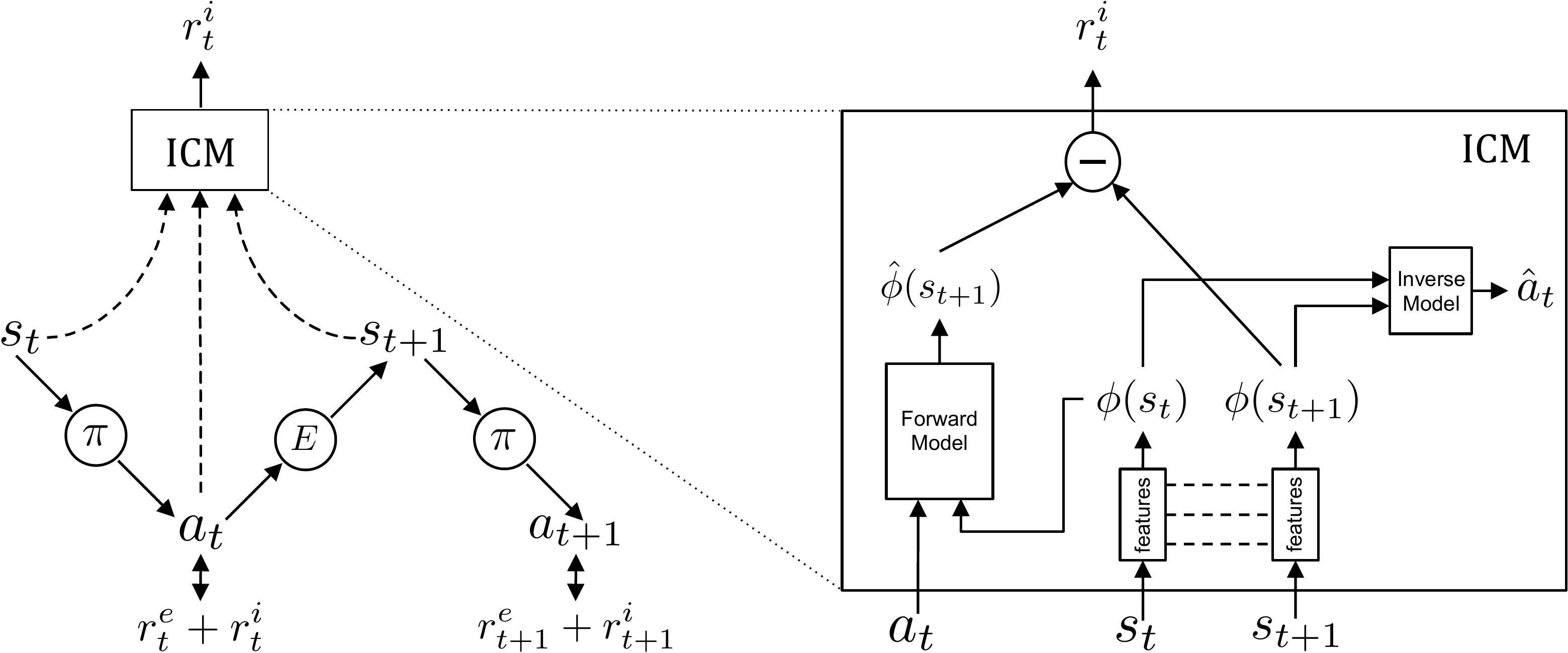}
\vspace{-.12in}
\caption{The agent in state $s_t$ interacts with the environment by executing an action $a_t$ sampled from its current policy $\pi$ and ends up in the state $s_{t+1}$. The policy $\pi$ is trained to optimize the sum of the extrinsic reward ($r_t^{e}$) provided by the environment \textit{E} and the curiosity based intrinsic reward signal ($r_t^{i}$) generated by our proposed Intrinsic Curiosity Module (ICM). ICM encodes the states $s_t$, $s_{t+1}$ into the features $\phi(s_t), \phi(s_{t+1})$ that are trained to predict $a_t$ (i.e. inverse dynamics model). The forward model takes as inputs $\phi(s_t)$ and $a_t$ and predicts the feature representation $\hat{\phi}(s_{t+1})$ of $s_{t+1}$. The prediction error in the feature space is used as the curiosity based intrinsic reward signal. As there is no incentive for $\phi(s_t)$ to encode any environmental features that can not influence or are not influenced by the agent's actions, the learned exploration strategy of our agent is robust to uncontrollable aspects of the environment.}
\label{fig:method}
\end{figure*}

\section{Curiosity-Driven Exploration}
\label{sec:method}
Our agent is composed of two subsystems: a reward generator that outputs a curiosity-driven intrinsic reward signal and a policy that outputs a sequence of actions to maximize that reward signal. In addition to intrinsic rewards, the agent optionally may also receive some extrinsic reward from the environment. Let the intrinsic curiosity reward generated by the agent at time $t$ be $r_t^i$ and the extrinsic reward be $r_t^e$. The policy sub-system is trained to maximize the sum of these two rewards $r_t = r_t^i + r_t^e$, with $r_t^e$ mostly (if not always) zero.

We represent the policy $\pi(s_t; \theta_P)$ by a deep neural network with parameters $\theta_P$. Given the agent in state $s_t$, it executes the action $a_t \sim \pi(s_t; \theta_P)$ sampled from the policy. $\theta_P$ is optimized to maximize the expected sum of rewards,
\begin{equation}
\label{eq:policy}
\max_{\theta_P} \mathbb{E}_{\pi(s_t; \theta_P)}[\Sigma_t r_t]
\end{equation}
Unless specified otherwise, we use the notation $\pi(s)$ to denote the parameterized policy $\pi(s; \theta_P)$.
Our curiosity reward model can potentially be used with a range of policy learning methods; in the experiments discussed here,
we use the asynchronous advantage actor critic policy gradient (A3C)~\cite{mnih2016asynchronous} for policy learning. Our main contribution is in designing an intrinsic reward signal based on prediction error of the agent's knowledge about its environment that scales to high-dimensional continuous state spaces like images, bypasses the hard problem of predicting pixels and is unaffected by the unpredictable aspects of the environment that do not affect the agent.

\subsection{Prediction error as curiosity reward}
\label{sec:pred_err}
Making predictions in the raw sensory space (e.g. when $s_t$ corresponds to images) is undesirable not only because it is hard to predict pixels directly, but also because it is unclear if predicting pixels is even the right objective to optimize. To see why, consider using prediction error in the pixel space as the curiosity reward. Imagine a scenario where the agent is observing the movement of tree leaves in a breeze. Since it is inherently hard to model breeze, it is even harder to predict the pixel location of each leaf.
This implies that the pixel prediction error will remain high and the agent will always remain curious about the leaves. But the motion of the leaves is inconsequential to the agent and therefore its continued curiosity about them is undesirable. The underlying problem is that the agent is unaware that some parts of the state space simply cannot be modeled and thus the agent can fall into an artificial curiosity trap and stall its exploration. Novelty-seeking exploration schemes that record the counts of visited states in a tabular form (or their extensions to continuous state spaces) also suffer from this issue. Measuring learning progress instead of prediction error has been proposed in the past as one solution~\cite{schmidhuber1991possibility}. Unfortunately, there are currently no known computationally feasible mechanisms for measuring learning progress.

If not the raw observation space, then what is the right feature space for making predictions so that the prediction error provides a good measure of curiosity? To answer this question, let us divide all sources that can modify the agent's observations into three cases: (1) things that can be controlled by the agent; (2) things that the agent cannot control but that can affect the agent (e.g. a vehicle driven by another agent), and (3) things out of the agent's control and not affecting the agent (e.g. moving leaves). A good feature space for curiosity should model (1) and (2) and be unaffected by (3). This latter is because, if there is a source of variation that is inconsequential for the agent, then the agent has no incentive to know about it.

\subsection{Self-supervised prediction for exploration}
Instead of hand-designing a feature representation for every environment, our aim is to come up with a general mechanism for learning feature representations such that the prediction error in the learned feature space provides a good intrinsic reward signal. We propose that such a feature space can be learned by training a deep neural network with two sub-modules: the first sub-module encodes the raw state $(s_t)$ into a feature vector $\phi(s_t)$ and the second sub-module takes as inputs the feature encoding $\phi(s_t), \phi(s_{t+1})$ of two consequent states and predicts the action $(a_t)$ taken by the agent to move from state $s_t$ to $s_{t+1}$. Training this neural network amounts to learning function $g$ defined as:
\begin{equation}
    \hat{a}_t = g\Big(s_t, s_{t+1}; \theta_I\Big)
\end{equation}
where, $\hat{a}_t$ is the predicted estimate of the action $a_t$ and the the neural network parameters $\theta_I$ are trained to optimize,
\begin{equation}
   \label{eq:loss_inverse}
   \min_{\theta_I}L_I(\hat{a}_t, a_t)
\end{equation}
where, $L_I$ is the loss function that measures the discrepancy between the predicted and actual actions. In case $a_t$ is discrete, the output of $g$ is a soft-max distribution across all possible actions and minimizing $L_I$ amounts to maximum likelihood estimation of $\theta_I$ under a multinomial distribution.  The learned function $g$ is also known as the inverse dynamics model and the tuple $(s_t, a_t, s_{t+1})$ required to learn $g$ is obtained while the agent interacts with the environment using its current policy $\pi(s)$.

In addition to inverse dynamics model, we train another neural network that takes as inputs $a_t$ and $\phi(s_t)$ and predicts the feature encoding of the state at time step $t+1$,
\begin{equation}
   \hat{\phi}(s_{t+1}) = f\Big(\phi(s_{t}), a_t; \theta_F\Big)
\end{equation}
where $\hat{\phi}(s_{t+1})$ is the predicted estimate of $\phi(s_{t+1})$ and the neural network parameters $\theta_F$ are optimized by minimizing the loss function $L_F$:
\begin{equation}
\label{eq:loss_forward}
L_F\Big(\phi(s_{t}), \hat{\phi}(s_{t+1})\Big) = \frac{1}{2} \| \hat{\phi}(s_{t+1}) - \phi(s_{t+1}) \|_2^{2}
\end{equation}
The learned function $f$ is also known as the forward dynamics model.
The intrinsic reward signal $r_t^{i}$ is computed as,
\begin{equation}
   r_t^{i} = \frac{\eta}{2} \| \hat{\phi}(s_{t+1}) - \phi(s_{t+1}) \|_2^{2}
\end{equation}
where $\eta > 0$ is a scaling factor. In order to generate the curiosity based intrinsic reward signal, we jointly optimize the forward and inverse dynamics loss described in equations \ref{eq:loss_inverse} and \ref{eq:loss_forward} respectively. The inverse model learns a feature space that encodes information relevant for predicting the agent's actions only and the forward model makes predictions in this feature space. We refer to this proposed curiosity formulation as Intrinsic Curiosity Module (ICM). As there is no incentive for this feature space to encode any environmental features that are not influenced by the agent's actions, our agent will receive no rewards for reaching environmental states that are inherently unpredictable and its exploration strategy will be robust to the presence of distractor objects, changes in illumination,  or other nuisance sources of variation in the environment. See Figure~\ref{fig:method} for illustration of the formulation.

The use of inverse models has been investigated to learn features for recognition tasks~\cite{jayaraman2015learning,agrawal2015learning}. Agrawal et al.~\yrcite{agrawal2016learning} constructed a joint inverse-forward model to learn feature representation for the task of pushing objects. However, they only used the forward model as a regularizer for training the inverse model features, while we make use of the error in the forward model predictions as the curiosity reward for training our agent's policy.

The overall optimization problem that is solved for learning the agent is a composition of equations \ref{eq:policy}, \ref{eq:loss_inverse} and \ref{eq:loss_forward} and can be written as,
\begin{equation}
   \min_{\theta_P, \theta_I, \theta_F}\Bigg[-\lambda \mathbb{E}_{\pi(s_t; \theta_P)}[\Sigma_t r_t] + (1 - \beta)L_I + \beta L_F \Bigg]
\end{equation}
where $0 \leq \beta \leq 1$ is a scalar that weighs the inverse model loss against the forward model loss and $\lambda > 0$ is a scalar that weighs the importance of the policy gradient loss against the importance of learning the intrinsic reward signal.

\section{Experimental Setup}
\label{sec:expsetup}
To evaluate our curiosity module on its ability to improve exploration and provide generalization to novel scenarios, we will use two simulated environments. This section describes the details of the environments and the experimental setup.

\begin{figure}[t]
\centering
\begin{subfigure}[b]{0.54\linewidth}
    \includegraphics[width=\linewidth]{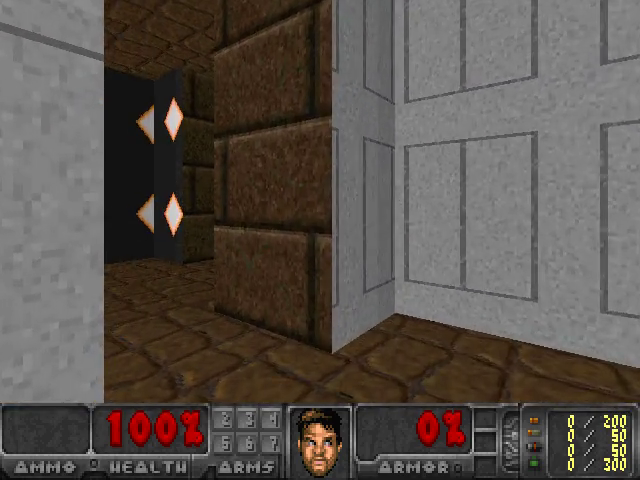}
    \vspace{-0.2in}
    \caption{Input snapshot in VizDoom}
    \label{fig:doom_snaps_a}
\end{subfigure}
~
\begin{subfigure}[b]{0.405\linewidth}
    \includegraphics[width=\linewidth]{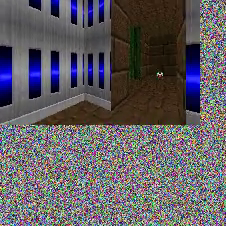}
    \vspace{-0.2in}
    \caption{Input w/ noise}
    \label{fig:doom_snaps_b}
\end{subfigure}
\vspace{-0.15in}
\caption{Frames from VizDoom 3-D environment which agent takes as input: (a) Usual 3-D navigation setup; (b) Setup when uncontrollable noise is added to the input.}
\label{fig:doom_snaps}
\end{figure}

\begin{figure}[t]
\centering
\begin{subfigure}[b]{0.44\linewidth}
    \includegraphics[width=\linewidth]{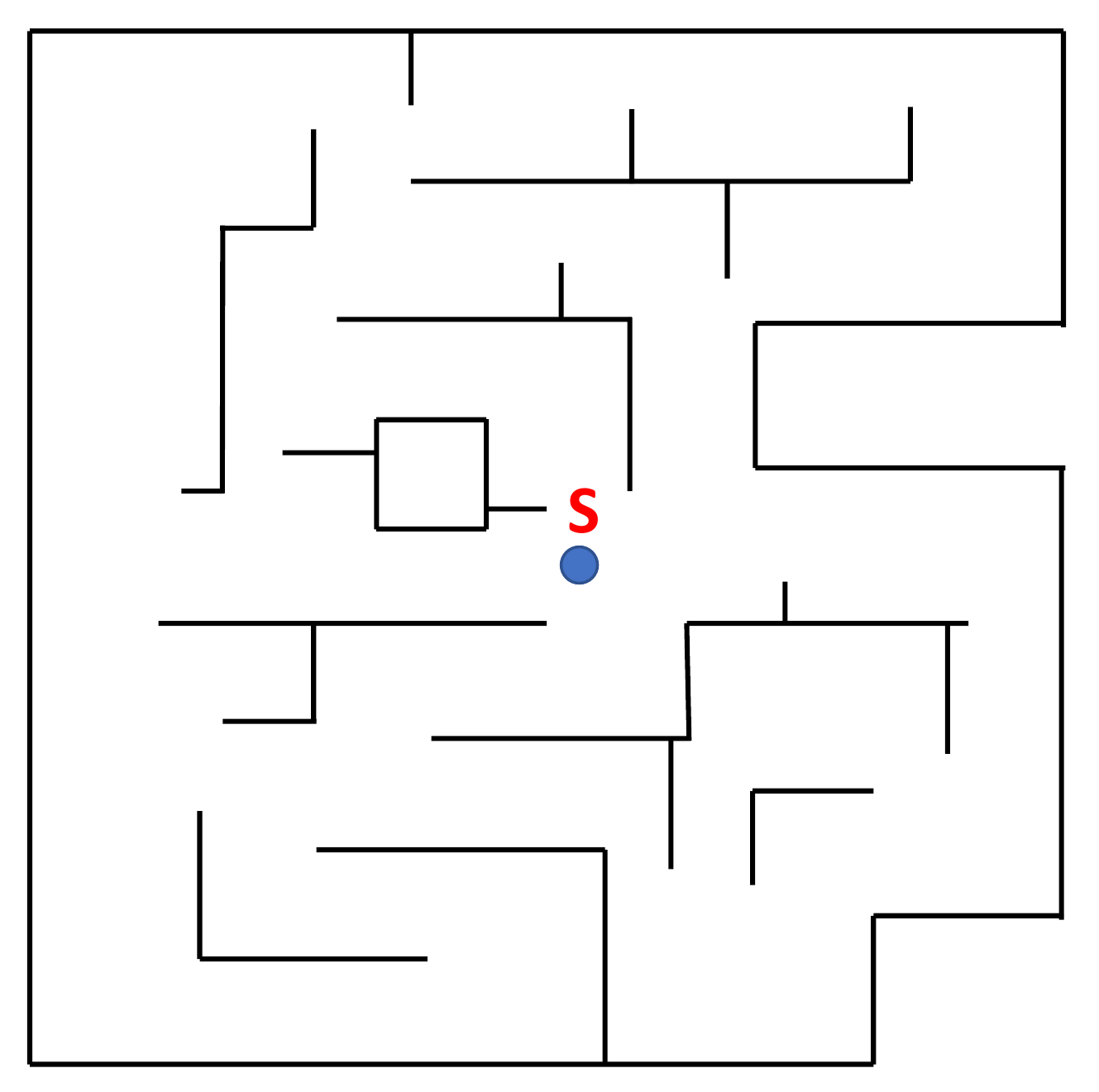}
    \vspace{-0.2in}
    \caption{Train Map Scenario}
    \label{fig:doom_map_a}
\end{subfigure}
~~
\begin{subfigure}[b]{0.50\linewidth}
    \includegraphics[width=\linewidth]{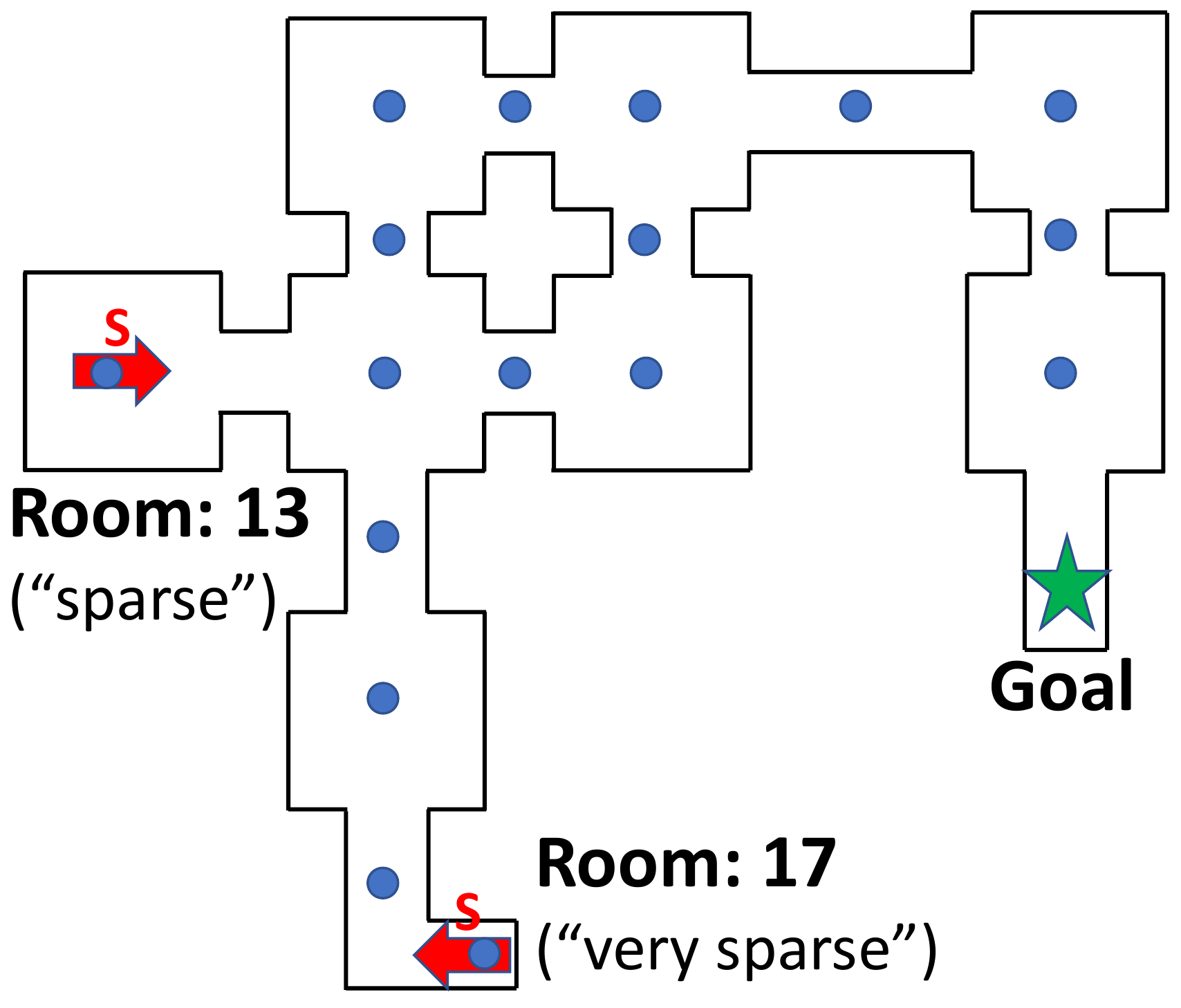}
    \vspace{-0.2in}
    \caption{Test Map Scenario}
    \label{fig:doom_map_b}
\end{subfigure}
\vspace{-0.15in}
\caption{Maps for VizDoom 3-D environment: (a) For generalization experiments (c.f. Section ~\ref{sec:generalization}), map of the environment where agent is pre-trained only using curiosity signal without any reward from environment. `S' denotes the starting position. (b) Testing map for VizDoom experiments. Green star denotes goal location. Blue dots refer to 17 agent spawning locations in the map in the ``dense'' case. Rooms 13, 17 are the fixed start locations of agent in ``sparse'' and ``very sparse'' reward cases respectively. Note that textures are also different in train and test maps.}
\label{fig:doom_map}
\vspace{-2pt}
\end{figure}

\paragraph{Environments}
The first environment we evaluate on is the VizDoom~\cite{doom} game. We consider the Doom 3-D navigation task where the action space of the agent consists of four discrete actions -- move forward, move left, move right and no-action.
Our testing setup in all the experiments is the `DoomMyWayHome-v0' environment which is available as part of OpenAI Gym~\cite{openaigym}.
Episodes are terminated either when the agent finds the vest or if the agent exceeds a maximum of 2100 time steps.
The map consists of $9$ rooms connected by corridors and the agent is tasked to reach some fixed goal location from its spawning location.
The agent is only provided a sparse terminal reward of +1 if it finds the vest and zero otherwise.
For generalization experiments, we pre-train on a different map with different random textures from~\cite{dosovitskiy2016learning} and each episode lasts for 2100 time steps.
Sample frames from VizDoom are shown in Figure~\ref{fig:doom_snaps_a}, and maps are explained in Figure~\ref{fig:doom_map}.
It takes approximately 350 steps for an optimal policy to reach the vest location from the farthest room in this map (sparse reward).

Our second environment is the classic Nintendo game Super Mario Bros~\cite{mariogym}.
We consider four levels of the game: pre-training on the first level and showing generalization on the subsequent levels.
In this setup, we reparametrize the action space of the agent into 14 unique actions following~\cite{mariogym}.
This game is played using a joystick allowing for multiple simultaneous button presses, where the duration of the press affects what action is being taken. This property makes the game particularly hard, e.g. to make a long jump over tall pipes or wide gaps, the agent needs to predict the same action up to 12 times in a row, introducing long-range dependencies.
All our experiments on Mario are trained using curiosity signal only, without any reward from the game.

\paragraph{Training details}
All agents in this work are trained using \textit{visual inputs} that are pre-processed in manner similar to \cite{mnih2016asynchronous}. The input RGB images are converted into gray-scale and re-sized to $42\times 42$.
In order to model temporal dependencies, the state representation ($s_t$) of the environment is constructed by concatenating the current frame with the three previous frames.
Closely following \cite{mnih2015human,mnih2016asynchronous}, we use action repeat of four during training time in VizDoom and action repeat of six in Mario.
However, we sample the policy without any action repeat during inference.
Following the asynchronous training protocol in A3C, all the agents were trained asynchronously with twenty workers using stochastic gradient descent.
We used ADAM optimizer with its parameters not shared across the workers.

\paragraph{A3C architecture}
The input state $s_t$ is passed through a sequence of four convolution layers with 32 filters each, kernel size of 3x3, stride of 2 and padding of 1. An exponential linear unit (ELU; \cite{clevert2015fast}) is used after each convolution layer. The output of the last convolution layer is fed into a LSTM with 256 units. Two seperate fully connected layers are used to predict the value function and the action from the LSTM feature representation.

\paragraph{Intrinsic Curiosity Module (ICM) architecture}
The intrinsic curiosity module consists of the forward and the inverse model. The inverse model first maps the input state $(s_t)$ into a feature vector $\phi(s_t)$ using a series of four convolution layers, each with 32 filters, kernel size 3x3, stride of 2 and padding of 1. ELU non-linearity is used after each convolution layer. The dimensionality of $\phi(s_t)$ (i.e. the output of the fourth convolution layer) is 288. For the inverse model, $\phi(s_t)$ and $\phi(s_{t+1})$ are concatenated into a single feature vector and passed as inputs into a fully connected layer of 256 units followed by an output fully connected layer with 4 units to predict one of the four possible actions. The forward model is constructed by concatenating $\phi(s_t)$ with $a_t$ and passing it into a sequence of two fully connected layers with 256 and 288 units respectively. The value of $\beta$ is $0.2$, and $\lambda$ is 0.1. The Equation (7) is minimized with learning rate of $1e-3$.

\begin{figure*}[t!]
\centering
\begin{subfigure}[b]{0.34\linewidth}
    \includegraphics[width=\linewidth]{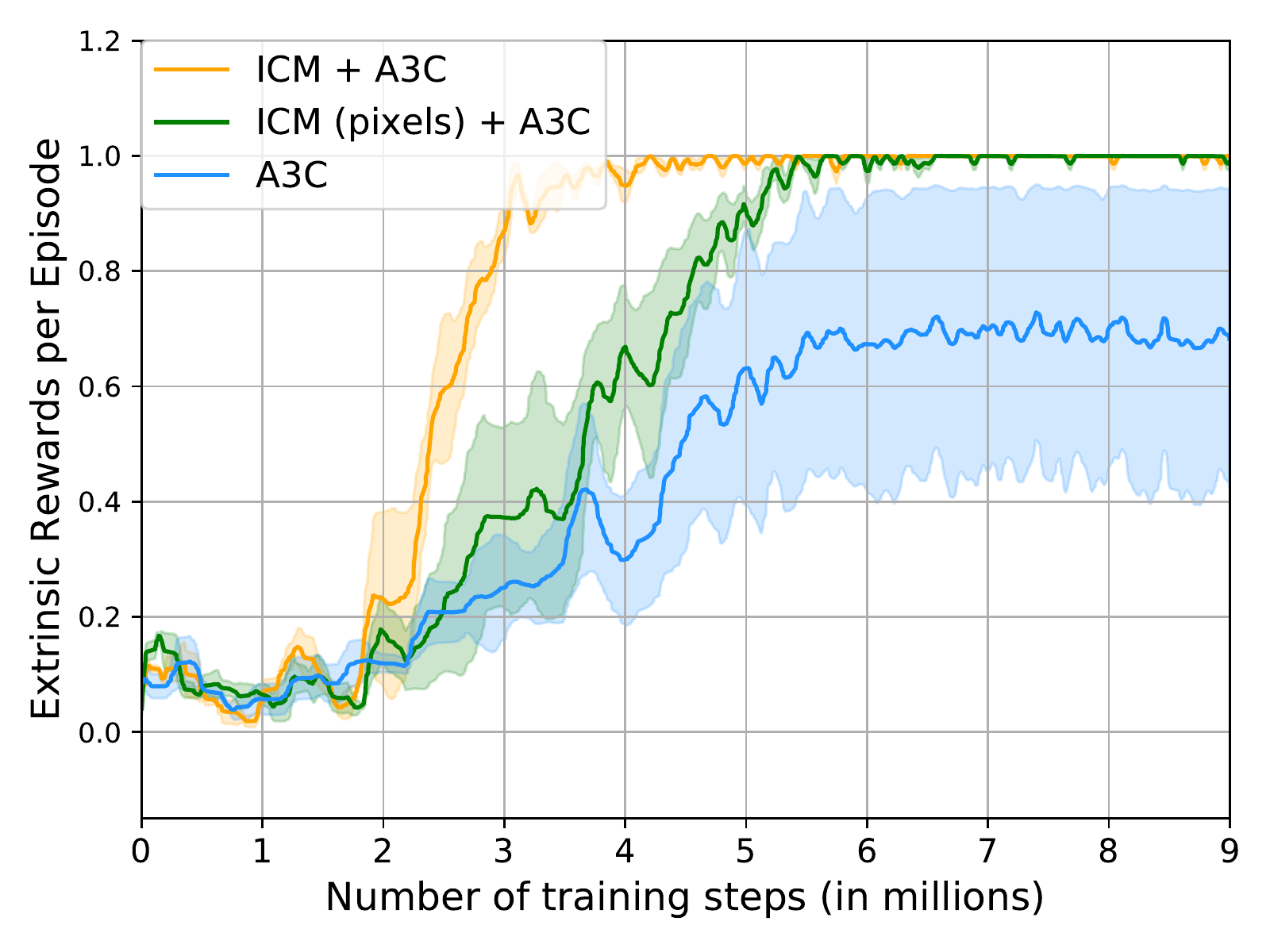}
    \vspace{-0.15in}
    \caption{``dense reward'' setting}
    \label{fig:doom_sparse_a}
\end{subfigure}
\begin{subfigure}[b]{0.32\linewidth}
    \includegraphics[width=\linewidth]{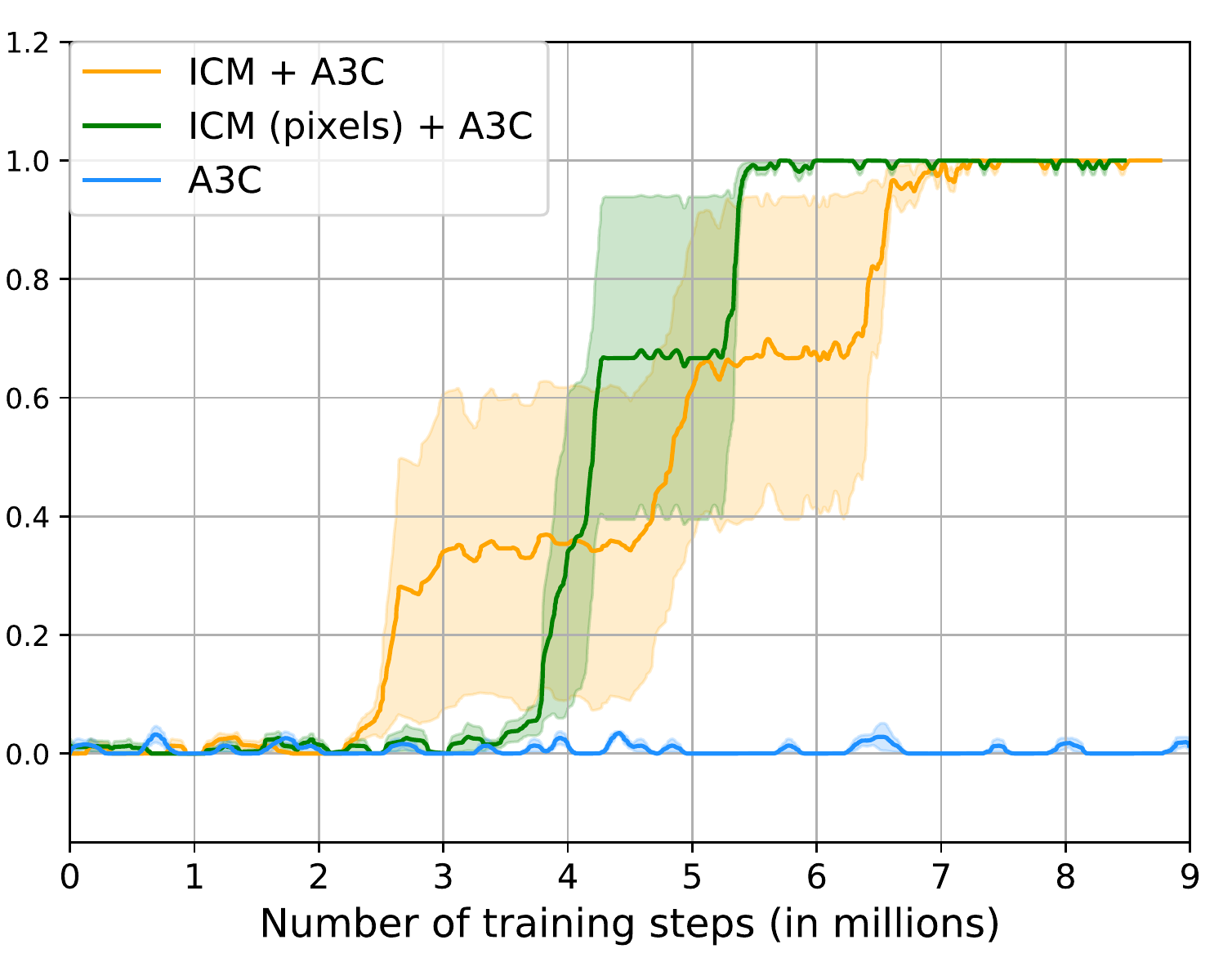}
    \vspace{-0.15in}
    \caption{``sparse reward'' setting}
    \label{fig:doom_sparse_b}
\end{subfigure}
\begin{subfigure}[b]{0.32\linewidth}
    \includegraphics[width=\linewidth]{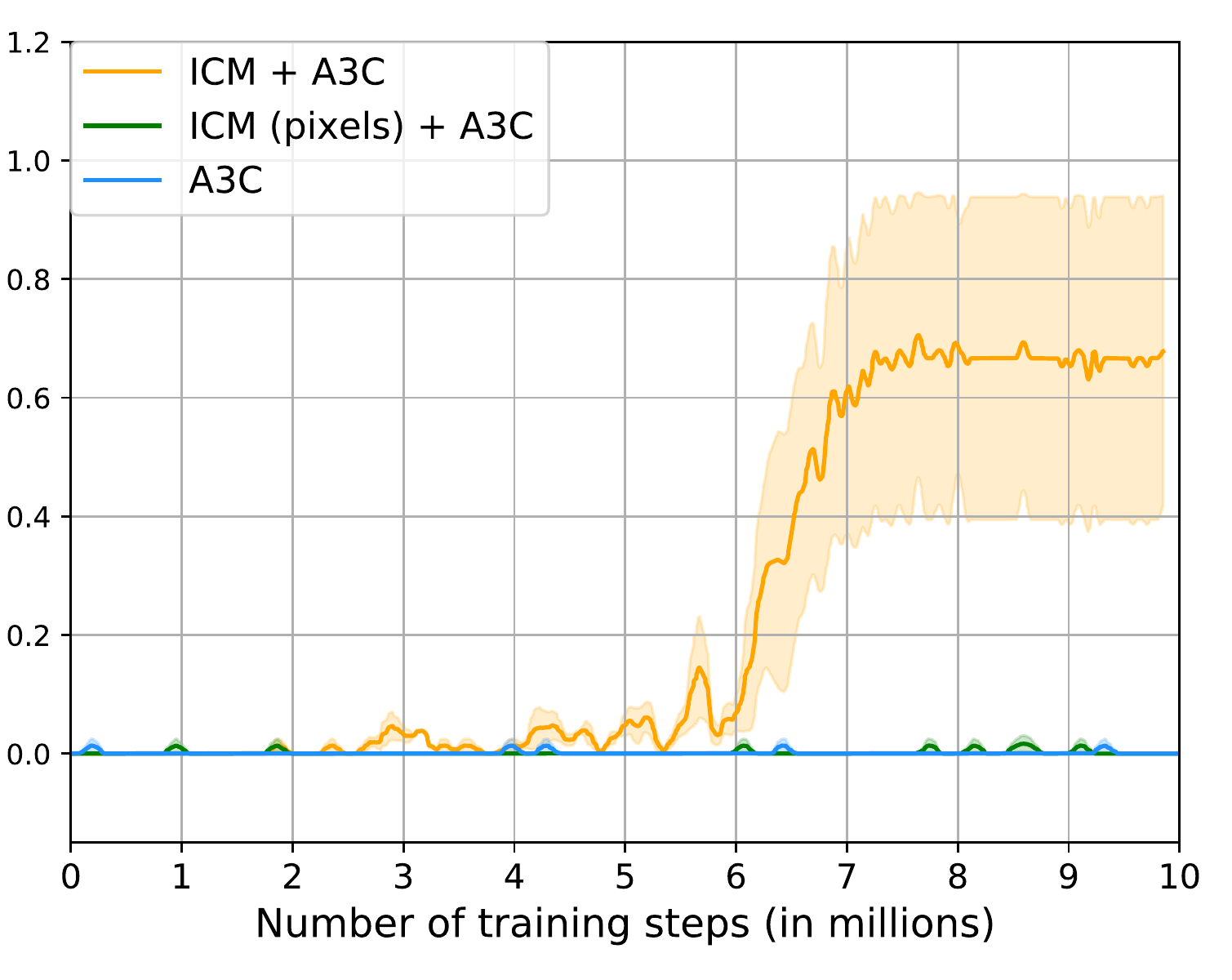}
    \vspace{-0.15in}
    \caption{``very sparse reward'' setting}
    \label{fig:doom_sparse_c}
\end{subfigure}
\vspace{-0.15in}
\caption{Comparing the performance of the A3C agent with no curiosity (blue) against the curiosity in pixel space agent (green) and the proposed curious ICM-A3C agent (orange) as the hardness of the exploration task is gradually increased from left to right. Exploration becomes harder with larger distance between the initial and goal locations: ``dense'', ``sparse'' and ``very sparse''. The results depict that succeeding on harder exploration task becomes progressively harder for the baseline A3C, whereas the curious A3C is able to achieve good score in all the scenarios. Pixel based curiosity works in dense and sparse but fails in very sparse reward setting. The protocol followed in the plots involves running three independent runs of each algorithm. Darker line represents mean and shaded area represents mean $\pm$ standard error of mean. We did not perform any tuning of random seeds.}
\label{fig:doom_sparse}
\end{figure*}

\paragraph{Baseline Methods}
`ICM + A3C' denotes our full algorithm which combines intrinsic curiosity model with A3C. Across different experiments, we compare our approach with three baselines. First is the vanilla `A3C' algorithm with $\epsilon$-greedy exploration. Second is `ICM-pixels + A3C', which is a variant of our ICM without the inverse model, and has curiosity reward dependent only on the forward model loss in predicting next observation in pixel space. To design this, we remove the inverse model layers and append deconvolution layers to the forward model. ICM-pixels is close to ICM in architecture but incapable of learning embedding that is invariant to the uncontrollable part of environment. Note that ICM-pixels is representative of previous methods which compute information gain by directly using the observation space~\cite{stadie2015incentivizing,schmidhuber2010formal}. We show that directly using observation space for computing curiosity is significantly worse than learning an embedding as in ICM. Finally, we include comparison with state-of-the-art exploration methods based on variational information maximization (VIME)~\cite{houthooft2016vime} which is trained with TRPO.

\section{Experiments}
\label{sec:results}
We qualitatively and quantitatively evaluate the performance of the learned policy with and without the proposed intrinsic curiosity signal in two environments, {\em VizDoom} and {\em Super Mario Bros}.
Three broad settings are evaluated:
a) sparse extrinsic reward on reaching a goal (Section~\ref{sec:sparse});
b) exploration with no extrinsic reward (Section~\ref{sec:noreward}); and c) generalization
to novel scenarios (Section~\ref{sec:generalization}).
In {\em VizDoom} generalization is evaluated on a novel map with novel textures, while in {\em Mario} it is evaluated on subsequent game levels.

\subsection{Sparse Extrinsic Reward Setting}
\label{sec:sparse}
We perform extrinsic reward experiments on {\em VizDoom} using `DoomMyWayHome-v0' setup described in Section~\ref{sec:expsetup}. The extrinsic reward is sparse and only provided when the agent finds the goal (a vest) located at a fixed location in the map. We systematically varied the difficulty of this goal-directed exploration task
by varying the distance between the initial spawning location of the agent and the location of the goal. A larger distance means that the chances of reaching the goal location by random exploration is lower and consequently the reward is said to be sparser.

\paragraph{Varying the degree of reward sparsity:} We consider three setups with ``dense'', ``sparse'' and ``very-sparse'' rewards (see Figure~\ref{fig:doom_map_b}). In these settings, the reward is always terminal and the episode terminates upon reaching goal or after a maximum of 2100 steps. In the ``dense'' reward case, the agent is randomly spawned in any of the 17 possible spawning locations uniformly distributed across the map. This is not a hard exploration task because sometimes the agent is randomly initialized close to the goal and therefore by random $\epsilon$-greedy exploration it can reach the goal with reasonably high probability. In the ``sparse'' and ``very sparse'' reward cases, the agent is always spawned in Room-13 and Room-17 respectively which are $~270$ and $~350$ steps away from the goal under an optimal policy. A long sequence of directed actions is required to reach the goals from these rooms, making these settings hard goal directed exploration problems.

\begin{figure}[t!]
\centering
\includegraphics[width=0.9\linewidth]{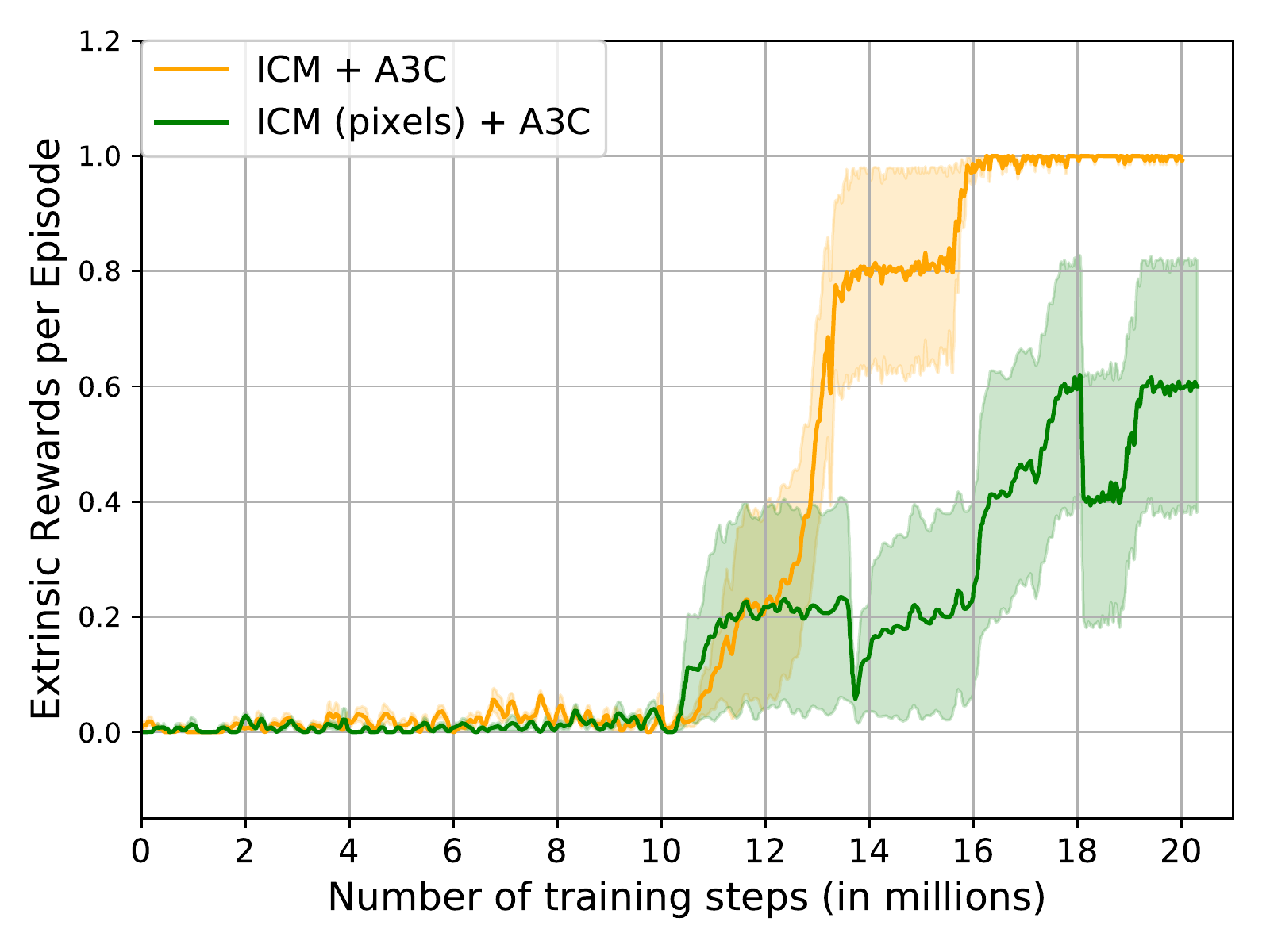}
\vspace{-0.15in}
\caption{Evaluating the robustness of ICM to the presence of uncontrollable distractors in the environment. We created such a distractor by replacing 40\% of the visual observation of the agent by white noise (see Figure~\ref{fig:doom_snaps_b}). The results show that while ICM succeeds most of the times, the pixel prediction model struggles.}
\label{fig:doom_noise}
\end{figure}

\begin{figure*}[t!]
\centering
\includegraphics[width=0.19\linewidth]{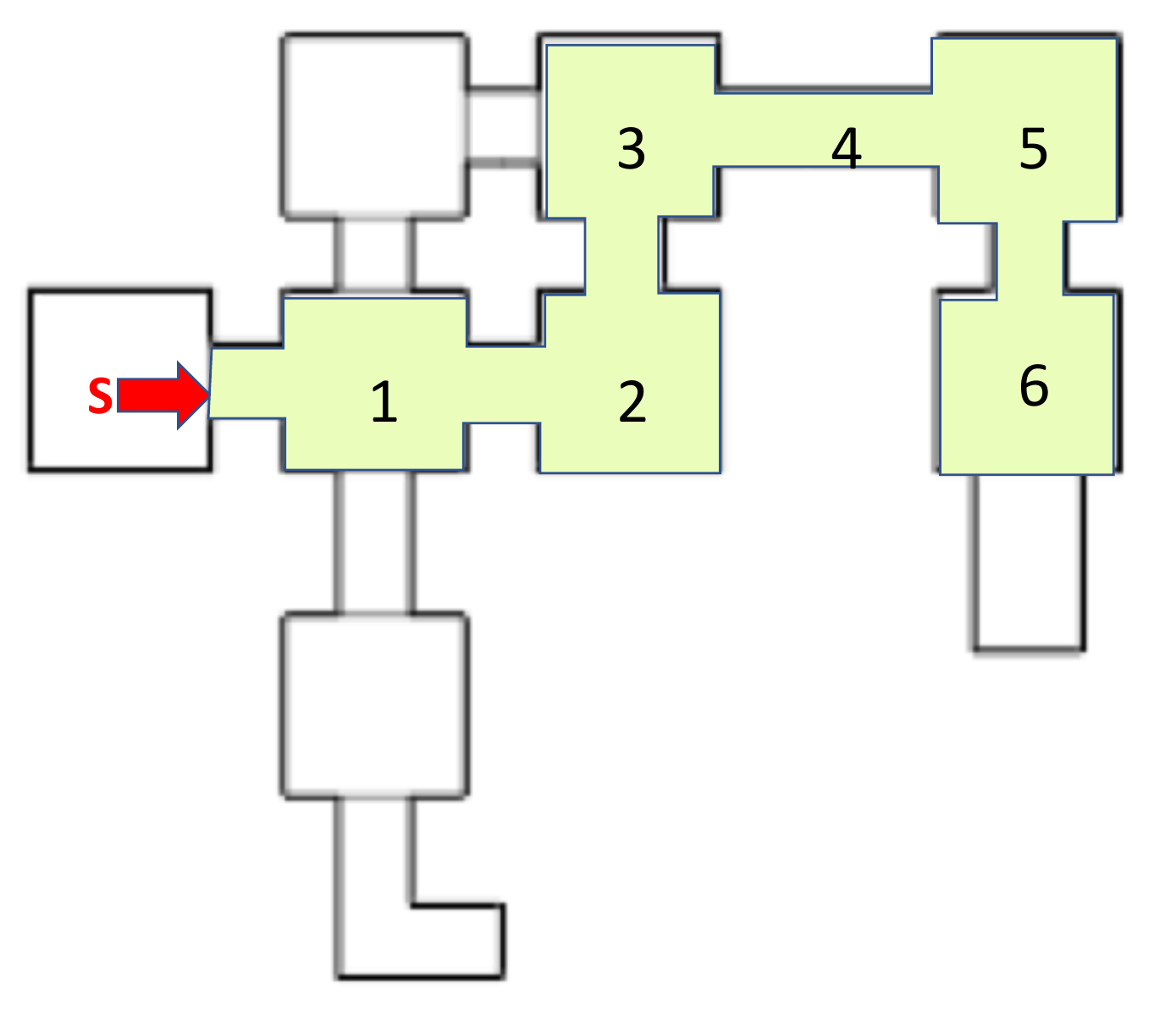}
\includegraphics[width=0.19\linewidth]{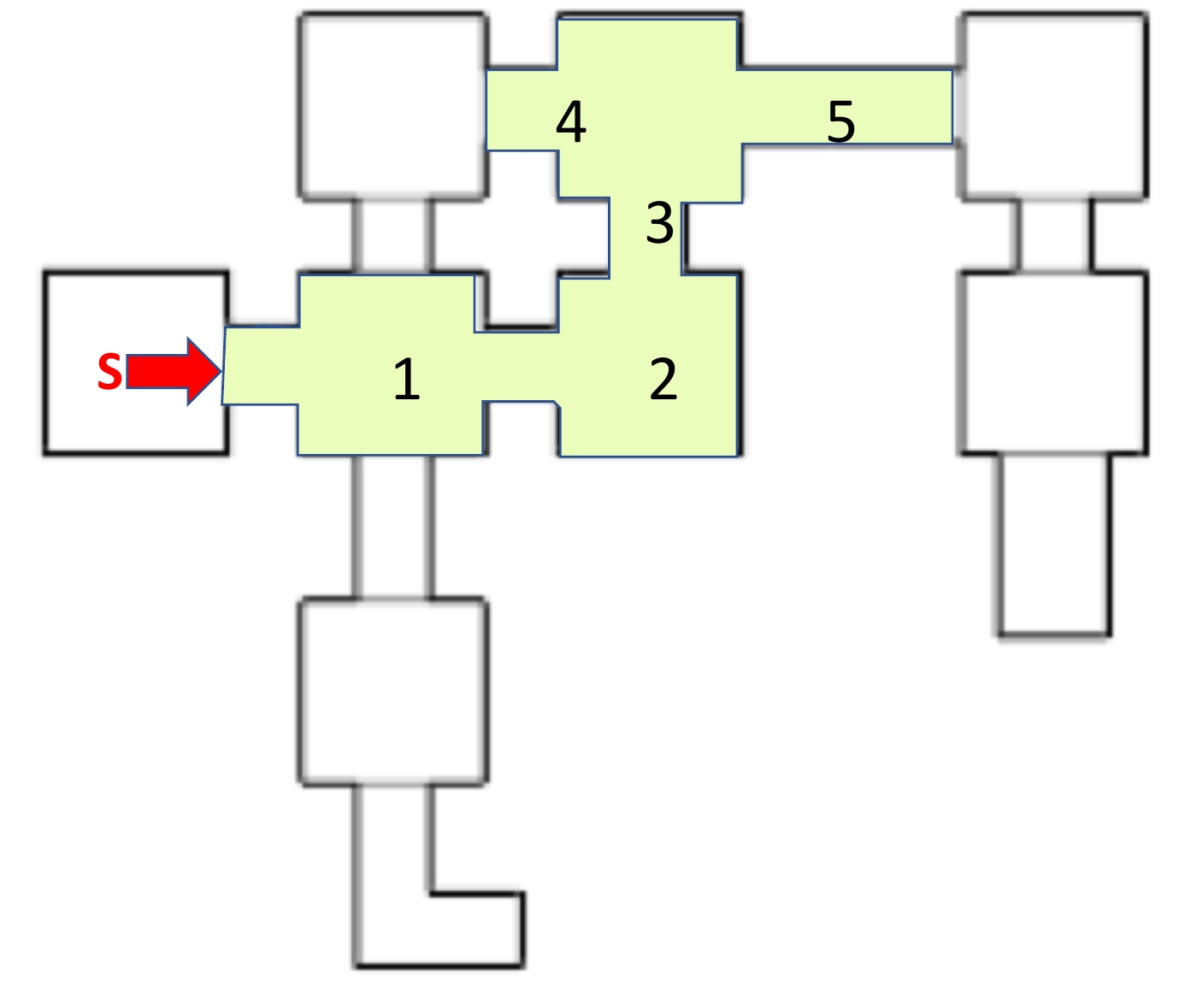}
\includegraphics[width=0.19\linewidth]{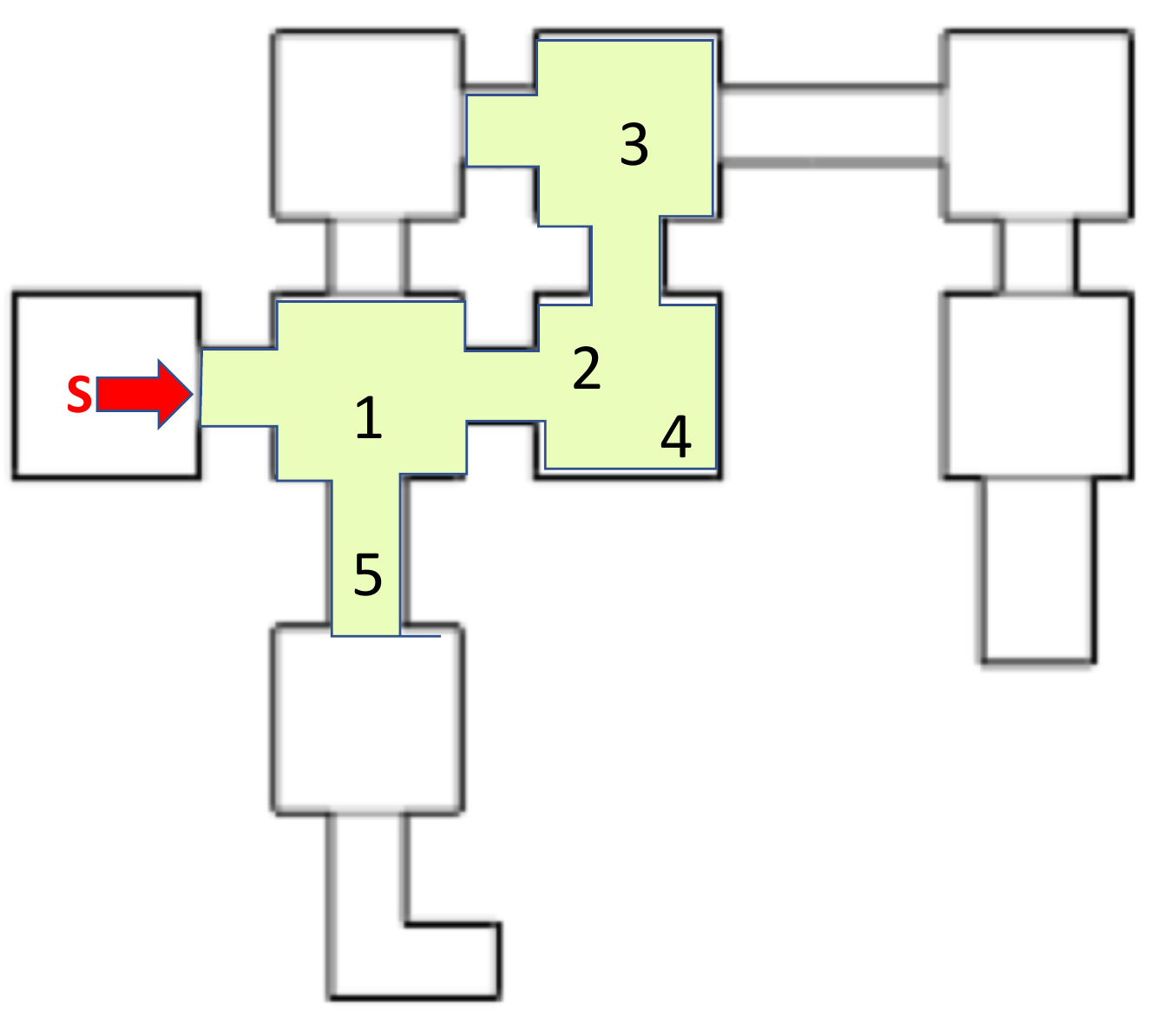}
\includegraphics[width=0.19\linewidth]{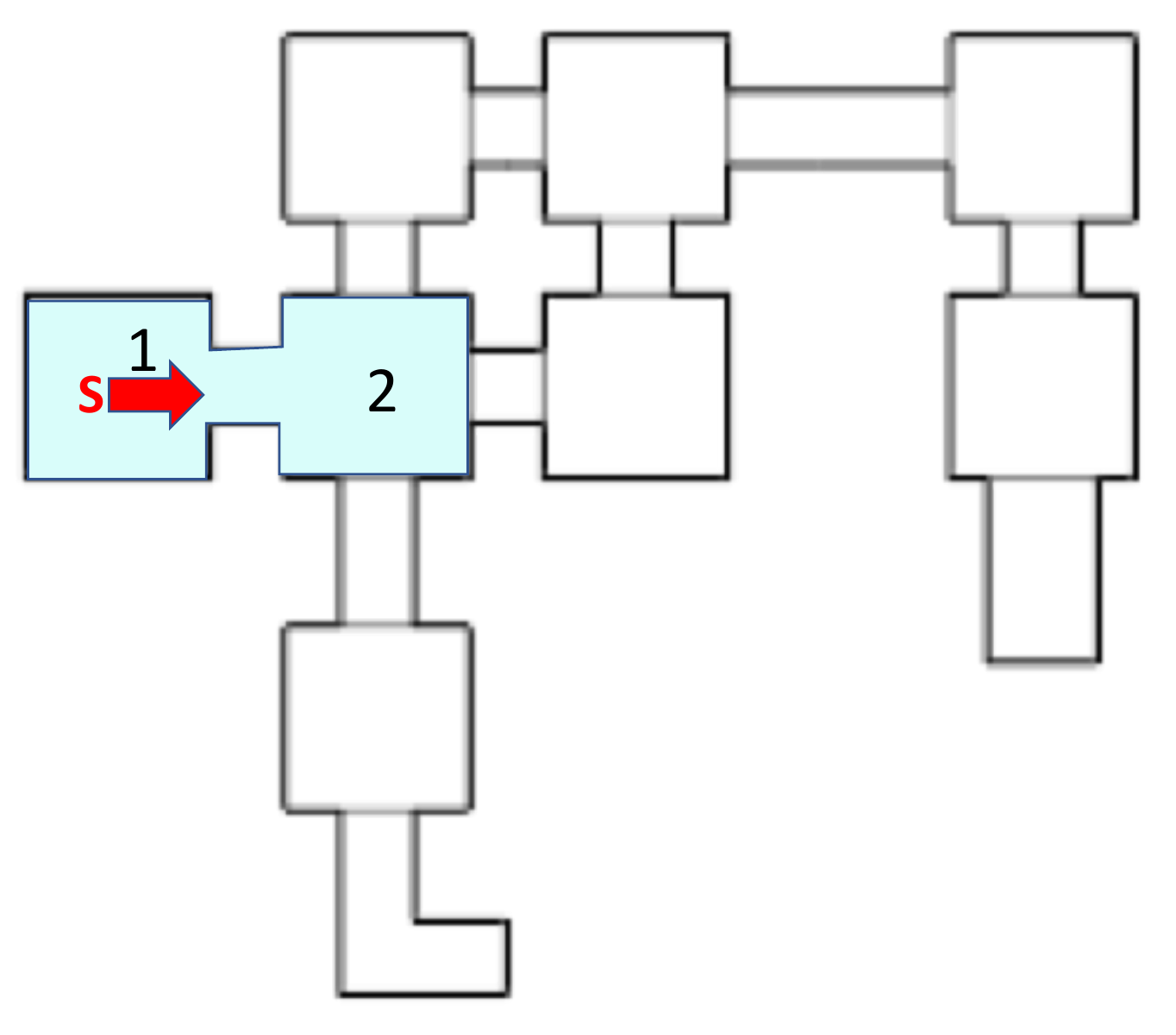}
\includegraphics[width=0.19\linewidth]{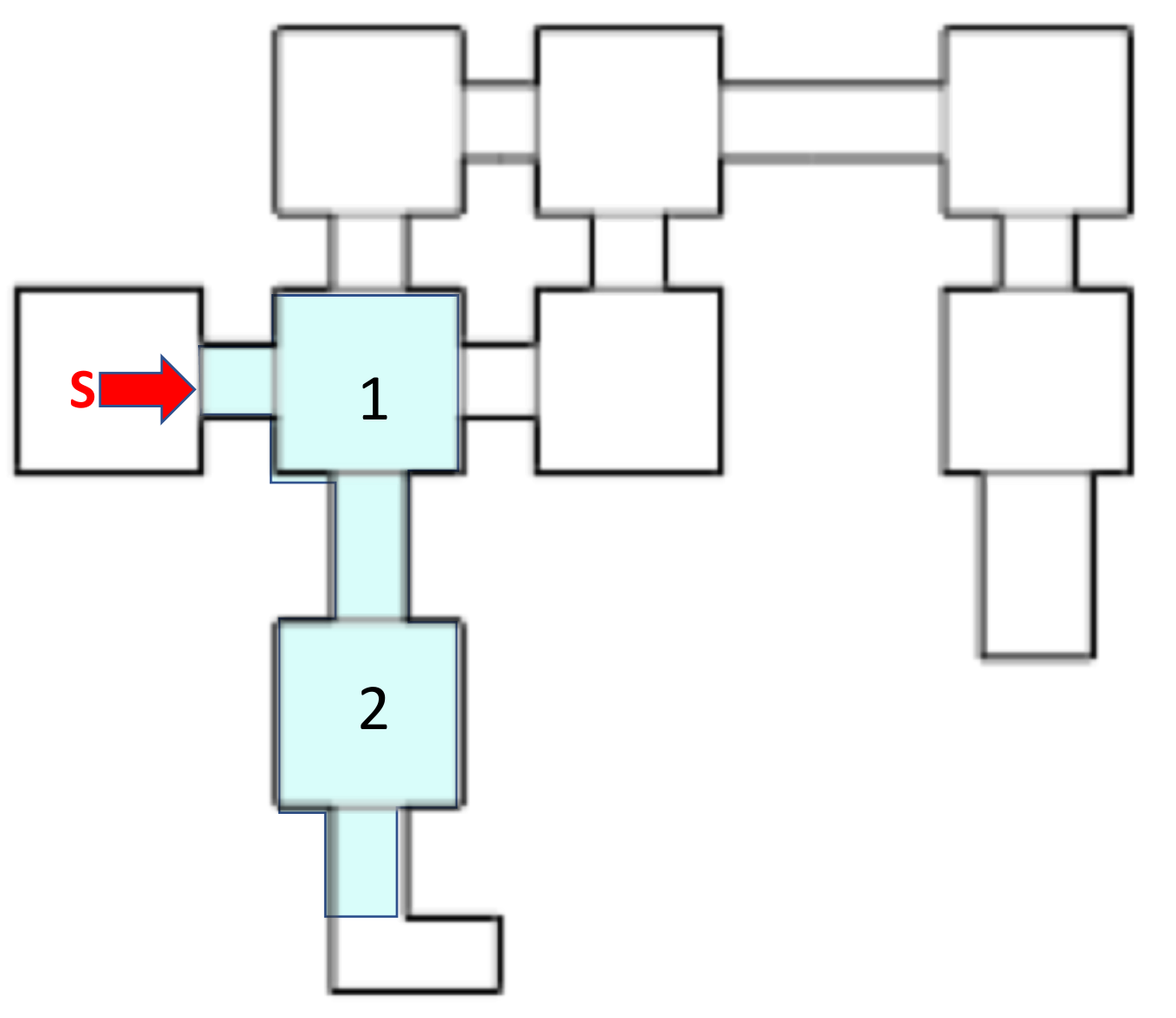}
\vspace{-0.15in}
\caption{Each column in the figure shows the visitation pattern of an agent exploring the environment. The red arrow shows the initial location and orientation of the agent at the start of the episode. Each room that the agent visits during its exploration of maximum 2100 steps has been colored. The first three columns (with maps colored in yellow) show the exploration strategy of an agent trained with curiosity driven internal reward signal only. The last two columns show the rooms visited by an agent conducting random exploration. The results clearly show that the curious agent trained with intrinsic rewards explores a significantly larger number of rooms as compared to a randomly exploring agent.}
\label{fig:no_rewards}
\end{figure*}

Results shown in Figure~\ref{fig:doom_sparse} indicate that while the performance of the baseline A3C degrades with sparser rewards, curious A3C agents are superior in all cases. In the ``dense'' reward case, curious agents learn much faster indicating more efficient exploration of the environment as compared to $\epsilon$-greedy exploration of the baseline agent. One possible explanation of the inferior performance of ICM-pixels in comparison to ICM is that in every episode the agent is spawned in one out of seventeen rooms with different textures. It is hard to learn a pixel-prediction model as the number of textures increases.

In the ``sparse'' reward case, as expected, the baseline A3C agent fails to solve the task, while the curious A3C agent is able to learn the task quickly. Note that ICM-pixels and ICM have similar convergence because, with a fixed spawning location of the agent, the ICM-pixels encounters the same textures at the starting of each episode which makes learning the pixel-prediction model easier as compared to the ``dense'' reward case. Finally, in the ``very sparse'' reward case, both the A3C agent and ICM-pixels never succeed, while the ICM agent achieves a perfect score in $66\%$ of the random runs. This indicates that ICM is better suited than ICM-pixels and vanilla A3C for hard goal directed exploration tasks.

\paragraph{Robustness to uncontrollable dynamics}
For testing the robustness of the proposed ICM formulation to changes in the environment that do not affect the agent, we augmented the agent's observation with a fixed region of white noise which made up $40\%$ of the image (see Figure~\ref{fig:doom_snaps_b}). In {\em VizDoom} 3-D navigation, ideally the agent should be unaffected by this noise as the noise does not affect the agent in anyway and is merely a nuisance. Figure~\ref{fig:doom_noise} compares the performance of ICM against some baseline methods on the ``sparse'' reward setup described above. While, the proposed ICM agent achieves a perfect score, ICM-pixels suffers significantly despite having succeeded at the ``sparse reward'' task when the inputs were not augmented with any noise (see Figure~\ref{fig:doom_sparse_b}). This indicates that in contrast to ICM-pixels, ICM is insensitive to nuisance changes in the environment.

\paragraph{Comparison to TRPO-VIME}
We now compare our curious agent against variational information maximization agent trained with TRPO~\cite{houthooft2016vime} for the VizDoom ``sparse'' reward setup described above.
TRPO is in general more sample efficient than A3C but takes a lot more wall-clock time.
We do not show these results in plots because TRPO and A3C have different setups.
The hyper-parameters and accuracy for the TRPO and VIME results follow from the concurrent work~\cite{fu2017ex2}.
Despite the sample efficiency of TRPO, we see that our ICM agents work significantly better than TRPO and TRPO-VIME, both in terms of convergence rate and accuracy.
Results are shown in the Table below:

\begin{center}
\footnotesize{
\begin{tabular}{cc}
\toprule
Method & Mean (Median) Score\\
(``sparse'' reward setup) & (at convergence)\\
\midrule
TRPO & \phantom{0}26.0 \% (\phantom{0}\phantom{0}0.0 \%)\\
A3C & \phantom{0}\phantom{0}0.0 \% (\phantom{0}\phantom{0}0.0 \%)\\
VIME + TRPO & \phantom{0}46.1 \% (\phantom{0}27.1 \%)\\
\midrule
ICM + A3C & \textbf{100.0} \% (100.0 \%)\\
\bottomrule
\label{tab:vime}
\end{tabular}
}
\end{center}

As a sanity check, we replaced the curiosity network with random noise sources and used them as the curiosity reward. We performed systematic sweep across different distribution parameters in the ``sparse'' reward case: uniform, Gaussian and Laplacian. We found that none of these agents were able to reach the goal showing that our curiosity module does not learn degenerate solutions.

\subsection{No Reward Setting}
\label{sec:noreward}
A good exploration policy is one which allows the agent to visit as many states as possible even without any goals. In the case of 3-D navigation, we expect a good exploration policy to cover as much of the map as possible; in the case of playing a game, we expect it to visit as many game states as possible. In order to test if our agent can learn a good exploration policy, we trained it on {\em VizDoom} and {\em Mario} without any rewards from the environment. We then evaluated what portion of the map was explore (for {\em VizDoom}), and how much progress it made (for {\em Mario}) in this setting.
To our surprise, we have found that in both cases, the no-reward agent was able to perform quote well (see video at \url{http://pathak22.github.io/noreward_rl/}).

{\bf VizDoom: Coverage during Exploration.}
An agent trained with no extrinsic rewards was able to learn to navigate corridors, walk between rooms and explore many rooms in the 3-D Doom environment. On many occasions the agent traversed the entire map and reached rooms that were farthest away from the room it was initialized in.
Given that the episode terminates in 2100 steps and farthest rooms are over 250 steps away (for an optimally-moving agent), this result is quite remarkable, demonstrating that it is possible to learn useful skills without the requirement of any external supervision of rewards.
Example explorations are shown in Figure~\ref{fig:no_rewards}. The first 3 maps show our agent explore a much larger state space without any extrinsic signal, compared to a random exploration agent (last two maps), which often has hard time getting around local minima of state spaces, e.g. getting stuck against a wall and not able to move (see video).

{\bf Mario: Learning to play with no rewards.}
We train our agent in the Super Mario World using only curiosity based signal.
Without any extrinsic reward from environment, our Mario agent can learn to cross over $30 \%$ of Level-1. The agent received no reward for killing or dodging enemies or avoiding fatal events, yet it automatically discovered these behaviors (see video). One possible reason is because getting killed by the enemy will result in only seeing a small part of the game space, making its curiosity saturate. In order to remain curious, it is in the agent's interest to learn how to kill and dodge enemies so that it can reach new parts of the game space. This suggests that curiosity provides indirect supervision for learning interesting behaviors.

To the best of our knowledge, this is the first demonstration where the agent learns to navigate in a 3D environment and discovers how to play a game by making use of relatively complex visual imagery directly from pixels, without any extrinsic rewards. There are several prior works that use reinforcement learning to navigate in 3D environments from pixel inputs or playing ATARI games such as~\cite{mnih2015human,mnih2016asynchronous,mirowski2016learning}, but they rely on intermediate external rewards provided by the environment.

\subsection{Generalization to Novel Scenarios}
\label{sec:generalization}
In the previous section we showed that our agent learns to explore large parts of the space where its curiosity-driven exploration policy was trained. However, it remains unclear whether the agent has done this by learning ``generalized skills'' for efficiently exploring its environment, or if it simply memorized the training set.
In other words we would like to know, when exploring a space, how much of the learned behavior is specific to that particular space and how much is general enough to be useful in novel scenarios? To investigate this question, we train a no reward exploratory behavior in one scenario (e.g. Level-1 of Mario) and then evaluate the resulting exploration policy in three different ways: a) apply the learned policy ``as is'' to a new scenario; b) adapt the policy by fine-tuning with curiosity reward only; c) adapt the policy to maximize some extrinsic reward. Happily, in all three cases, we observe some promising generalization results:

\begin{table*}[t]
\centering
\resizebox{\linewidth}{!}{%
\begin{tabular}{l|c|cccc|cccc}
\toprule
Level Ids & Level-1 & \multicolumn{4}{c|}{Level-2} & \multicolumn{4}{c}{Level-3} \\
\midrule
\small Accuracy &\small Scratch &\small Run as is &\small Fine-tuned &\small Scratch &\small Scratch &\small Run as is &\small Fine-tuned &\small Scratch &\small Scratch\\
\small Iterations &\small 1.5M &\small 0 &\small 1.5M &\small 1.5M &\small 3.5M &\small 0 &\small 1.5M &\small 1.5M &\small 5.0M\\
\midrule
\small Mean $\pm$ stderr & 711 $\pm$ 59.3 & 31.9 $\pm$ 4.2 & 466 $\pm$ 37.9 & 399.7 $\pm$ 22.5 & 455.5 $\pm$ 33.4 & 319.3 $\pm$ 9.7 & 97.5 $\pm$ 17.4 & 11.8 $\pm$ 3.3 & 42.2 $\pm$ 6.4\\
\small $\%$ distance $>$ 200 & 50.0 $\pm$ 0.0 & 0 & 64.2 $\pm$ 5.6 & 88.2 $\pm$ 3.3 & 69.6 $\pm$ 5.7 & 50.0 $\pm$ 0.0 & 1.5 $\pm$ 1.4 & 0 & 0\\
\small $\%$ distance $>$ 400 & 35.0 $\pm$ 4.1 & 0 & 63.6 $\pm$ 6.6 & 33.2 $\pm$ 7.1 & 51.9 $\pm$ 5.7 & 8.4 $\pm$ 2.8 & 0 & 0 & 0\\
\small $\%$ distance $>$ 600 & 35.8 $\pm$ 4.5 & 0 & 42.6 $\pm$ 6.1 & 14.9 $\pm$ 4.4 & 28.1 $\pm$ 5.4 & 0 & 0 & 0 & 0\\
\bottomrule
\end{tabular}}
\vspace{-0.15in}
\caption{Quantitative evaluation of the agent trained to play Super Mario Bros. using only curiosity signal without any rewards from the game. Our agent was trained with no rewards in Level-1. We then evaluate the agent's policy both when it is run ``as is'', and further fine-tuned on subsequent levels. The results are compared to settings when Mario agent is train from scratch in Level-2,3 using only curiosity without any extrinsic rewards. Evaluation metric is based on the distance covered by the Mario agent.}
\label{tab:mario_table}
\end{table*}

\paragraph{Evaluate ``as is'':}
We evaluate the policy trained by maximizing curiosity on Level-1 of {\em Mario} on subsequent levels without adapting the learned policy in any way. We measure the distance covered by the agent as a result of executing this policy on Levels 1, 2, and 3, as shown in Table~\ref{tab:mario_table}. We note that the policy performs surprisingly well on Level 3, suggesting good generalization, despite the fact that Level-3 has different structures and enemies compared to Level-1. However, note that the running ``as is'' on Level-2 does not do well.
At first, this seems to contradict the generalization results on Level-3. However, note that Level-3 has similar global visual appearance (day world with sunlight) to Level-1, whereas Level-2 is significantly different (night world). If this is indeed the issue, then it should be possible to quickly adapt the exploration policy to Level-2 with a little bit of ``fine-tuning''. We will explore this below.

\paragraph{Fine-tuning with curiosity only:}
From Table~\ref{tab:mario_table} we see that when the agent pre-trained (using only curiosity as reward) on Level-1 is fine-tuned (using only curiosity as reward) on Level-2 it quickly overcomes the mismatch in global visual appearance and achieves a higher score than training from scratch with the same number of iterations. Interestingly, training ``from scratch'' on Level-2
is worse than the fine-tuned policy, even when training for more iterations than pre-training + fine-tuning combined. One possible reason is that Level-2 is more difficult than Level-1, so learning the basic skills such as moving, jumping, and killing enemies from scratch is much more dangerous than in the relative ``safety'' of Level-1.
This result, therefore might suggest that first pre-training on an earlier level and then fine-tuning on a later one produces a form of curriculum which aids learning and generalization. In other words, the agent is able to use the knowledge it acquired by playing Level-1 to better explore the subsequent levels. Of course, the game designers do this on purpose to allow the human players to gradually learn to play the game.

However, interestingly, fine-tuning the exploration policy pre-trained on Level-1 to Level-3 deteriorates the performance,
compared to running ``as is''. This is because Level-3 is very hard for the agent to cross beyond a certain point -- the agent hits a curiosity blockade and is unable to make any progress. As the agent has already learned about parts of the environment before the hard point, it receives almost no curiosity reward and as a result it attempts to update its policy with almost zero intrinsic rewards and the policy slowly degenerates. This behavior is vaguely analogous to boredom, where if the agent is unable to make progress it gets bored and stops exploring.

\begin{figure}[t]
\centering
\includegraphics[width=0.9\linewidth]{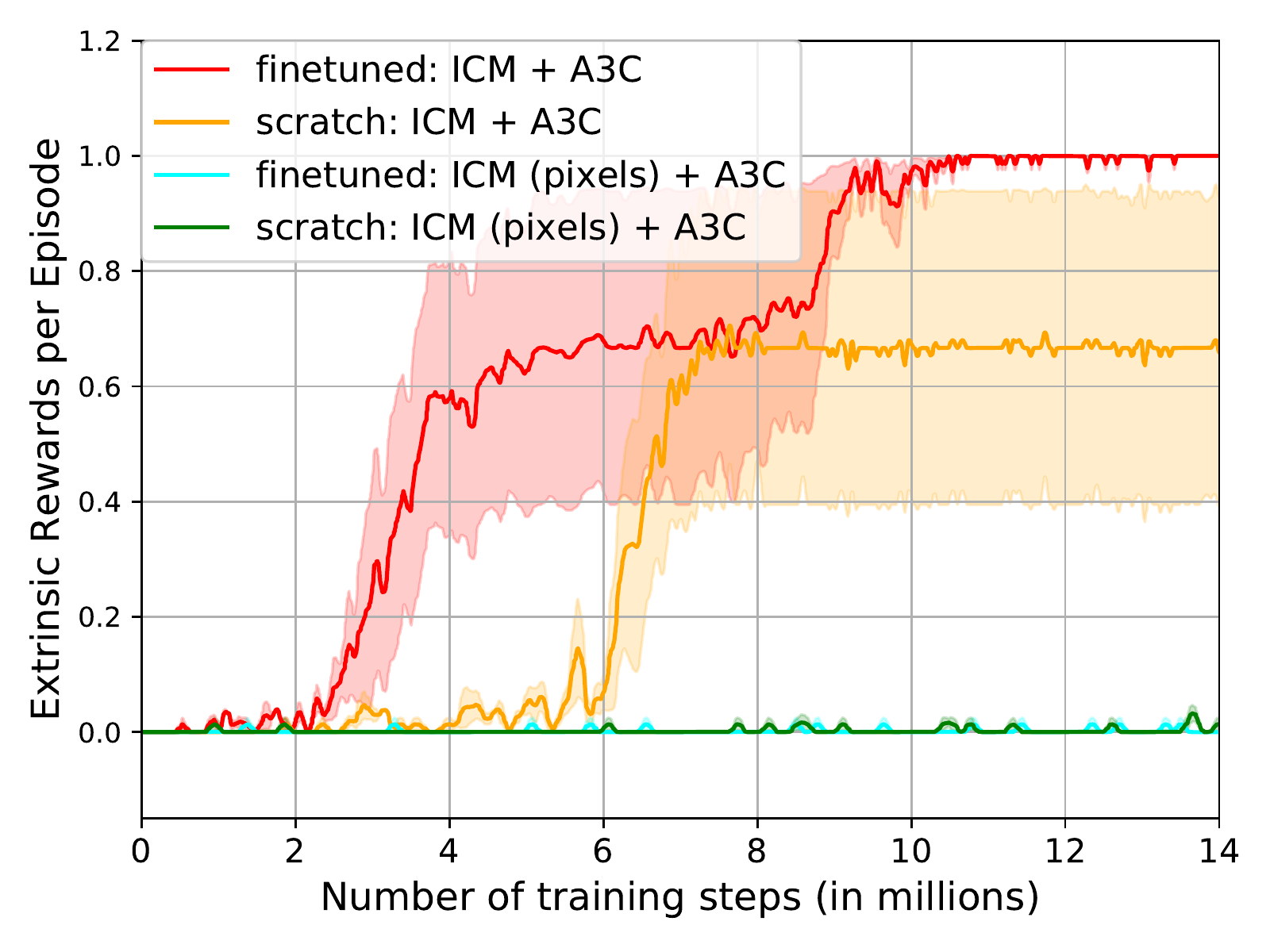}
\vspace{-0.15in}
\caption{Performance of ICM + A3C agents on the test set of {\em VizDoom} in the ``very sparse'' reward case. Fine-tuned models learn the exploration policy without any external rewards on the training maps and are then fine-tuned on the test map. The scratch models are directly trained on the test map. The fine-tuned ICM + A3C significantly outperforms ICM + A3C indicating that our curiosity formulation is able to learn generalizable exploration policies. The pixel prediction based ICM agent completely fail. Note that textures are also different in train and test.}
\label{fig:doom_gen}
\end{figure}

\paragraph{Fine-tuning with extrinsic rewards:}
If it is the case that the agent has actually learned useful exploratory behavior, then it should be able to learn quicker than starting from scratch even when external rewards are provided by environment. We perform this evaluation on {\em VizDoom} where we pre-train the agent using curiosity reward on a map showed in Figure~\ref{fig:doom_map_a}. We then test on the ``very sparse'' reward setting of `DoomMyWayHome-v0' environment which uses a different map with novel textures (see Figure~\ref{fig:doom_map_b}) as described earlier in Section \ref{sec:sparse}.

Results in Figure~\ref{fig:doom_gen} show that the ICM agent pre-trained only with curiosity and then fine-tuned with external reward learns faster and achieves higher reward than an ICM agent trained from scratch to jointly maximize curiosity and the external rewards. This result confirms that the learned exploratory behavior is also useful when the agent is required to achieve goals specified by the environment. It is also worth noting that ICM-pixels does not generalize to this test environment. This indicates that the proposed mechanism of measuring curiosity is significantly better for learning skills that generalize as compared to measuring curiosity in the raw sensory space.

\section{Related Work}
Curiosity-driven exploration is a well studied topic in the reinforcement learning literature and a good summary can be found in~\cite{oudeyer2007intrinsic,oudeyer2009intrinsic}. Schmidhuber ~\yrcite{schmidhuber1991possibility,schmidhuber2010formal} and Sun et al.~\yrcite{sun2011planning} use surprise and compression progress as intrinsic rewards. Classic work of Kearns et al.~\yrcite{kearns1999efficient} and Brafman et al.~\yrcite{brafman2002r} propose exploration algorithms polynomial in the number of state space parameters. Others have used empowerment, which is the information gain based on entropy of actions, as intrinsic rewards~\cite{klyubin2005empowerment,mohamed2015variational}. Stadie et al.~\yrcite{stadie2015incentivizing} use prediction error in the feature space of an auto-encoder as a measure of interesting states to explore.
State visitation counts have also been investigated for exploration~\cite{oh2015action,bellemare2016unifying,tang2016exploration}.
Osband et al.~\yrcite{osband2016deep} train multiple value functions and makes use of bootstrapping and Thompson sampling for exploration. Many approaches measure information gain for exploration~\cite{still2012information,little2014learning,storck1995reinforcement}. Houthooft et al.~\yrcite{houthooft2016vime} use an exploration strategy that maximizes information gain about the agent's belief of the environment's dynamics. Our approach of jointly training forward and inverse models for learning a feature space has similarities to~\cite{jordan1992forward,wolpert1995internal,agrawal2016learning}, but these works use the learned models of dynamics for planning a sequence of actions instead of exploration. The idea of using a proxy task to learn a semantic feature embedding has been used in a number of works on self-supervised learning in computer vision~\cite{goroshin2015unsupervised,agrawal2015learning,doersch2015unsupervised,jayaraman2015learning,wang2015unsupervised,pathak2016context}.

{\bf Concurrent work:}
A number of interesting related papers have appeared on Arxiv while the present work was in submission.
Sukhbaatar et al.~\yrcite{sukhbaatar2017intrinsic} generates supervision for pre-training via asymmetric self-play between two agents to improve data efficiency during fine-tuning. Several methods propose improving data efficiency of RL algorithms using self-supervised prediction based auxiliary tasks~\cite{jaderberg2017reinforcement,shelhamer2017loss}. Fu et al.~\yrcite{fu2017ex2} learn discriminative models, and Gregor et al.~\yrcite{gregor2017variational} use empowerment based measure to tackle exploration in sparse reward setups.

\section{Discussion}
\label{sec:discussion}
In this work we propose a mechanism for generating curiosity-driven intrinsic reward signal that scales to high dimensional visual inputs, bypasses the difficult problem of predicting pixels and ensures that the exploration strategy of the agent is unaffected by nuisance factors in the environment. We demonstrate that our agent significantly outperforms the baseline A3C with no curiosity, a recently proposed VIME~\cite{houthooft2016vime} formulation for exploration, and a baseline pixel-predicting formulation.

In {\em VizDoom} our agent learns the exploration behavior of moving along corridors and across rooms without any rewards from the environment. In {\em Mario} our agent crosses more than 30\% of Level-1 without any rewards from the game. One reason why our agent is unable to go beyond this limit is the presence of a pit at 38\% of the game that requires a very specific sequence of 15-20 key presses in order to jump across it. If the agent is unable to execute this sequence, it falls in the pit and dies, receiving no further rewards from the environment. Therefore it receives no gradient information indicating that there is a world beyond the pit that could potentially be explored. This issue is somewhat orthogonal to developing models of curiosity, but presents a challenging problem for policy learning.

It is common practice to evaluate reinforcement learning approaches in the same environment that was used for training. However, we feel that it is also important to evaluate on a separate ``testing set'' as well. This allows us to gauge how much of what has been learned is specific to the training environment (i.e. memorized), and how much might constitute ``generalizable skills'' that could be applied to new settings. In this paper, we evaluate generalization in two ways: 1) by applying the learned policy to a new scenario ``as is'' (no further learning), and 2) by fine-tuning the learned policy on a new scenario (we borrow the pre-training/fine-tuning nomenclature from the deep feature learning literature). We believe that evaluating generalization is a valuable tool and will allow the community to better understand the performance of various reinforcement learning algorithms. To further aid in this effort, we will make the code for our algorithm, as well as testing and environment setups freely available online.

An interesting direction of future research is to use the learned exploration behavior/skill as a motor primitive/low-level policy in a more complex, hierarchical system. For example, our {\em VizDoom} agent learns to walk along corridors instead of bumping into walls. This could be a useful primitive for a navigation system.

While the rich and diverse real world provides ample opportunities for interaction, reward signals are sparse. Our approach excels in this setting and converts unexpected interactions that affect the agent into intrinsic rewards. However our approach does not directly extend to the scenarios where ``opportunities for interactions'' are also rare. In theory, one could save such events in a replay memory and use them to guide exploration. However, we leave this extension for future work.

\paragraph{Acknowledgements:}
We would like to thank Sergey Levine, Evan Shelhamer, Saurabh Gupta, Phillip Isola and other members of the BAIR lab for fruitful discussions and comments. We thank Jacob Huh for help with Figure~\ref{fig:method} and Alexey Dosovitskiy for VizDoom maps. This work was supported in part by NSF IIS-1212798, IIS-1427425, IIS-1536003, IIS-1633310, ONR MURI N00014-14-1-0671, Berkeley DeepDrive, equipment grant from Nvidia, and the Valrhona Reinforcement Learning Fellowship.

\bibliography{main}
\bibliographystyle{icml2017}
\end{document}